\documentclass[runningheads]{llncs}
\usepackage{epsfig}
\usepackage{subfigure}
\usepackage{graphicx}
\usepackage{amsmath,amssymb} % define this before the line numbering.
\usepackage{bm}
\usepackage{color}
\usepackage{booktabs}
\usepackage{multirow}
\usepackage[ruled]{algorithm2e}
\usepackage[width=122mm,left=12mm,paperwidth=146mm,height=193mm,top=12mm,paperheight=217mm]{geometry}
\begin{document}
% \linenumbers
\pagestyle{headings}
\mainmatter
\def\ECCV18SubNumber{***}  % Insert your submission number here

\title{Novel View Synthesis for Large-scale Scene using Adversarial Loss} % Replace with your title

% \titlerunning{ECCV-18 submission ID \ECCV18SubNumber}

% \authorrunning{ECCV-18 submission ID \ECCV18SubNumber}

\author{Xiaochuan Yin, Henglai Wei, Penghong Lin, Xiangwei Wang, Qijun Chen}
% \institute{Paper ID \ECCV18SubNumber}

\maketitle

\begin{abstract}
Novel view synthesis aims to synthesize new images from different viewpoints of given images. Most of previous works focus on generating novel views of certain objects with a fixed background. However, for some applications, such as virtual reality or robotic manipulations, large changes in background may occur due to the egomotion of the camera. Generated images of a large-scale environment from novel views may be distorted if the structure of the environment is not considered. In this work, we propose a novel fully convolutional network, that can take advantage of the structural information explicitly by incorporating the inverse depth features. The inverse depth features are obtained from CNNs trained with sparse labeled depth values. This framework can easily fuse multiple images from different viewpoints. To fill the missing textures in the generated image, adversarial loss is applied, which can also improve the overall image quality. Our method is evaluated on the KITTI dataset. The results show that our method can generate novel views of large-scale scene without distortion. The effectiveness of our approach is demonstrated through qualitative and quantitative evaluation.
\dots
\keywords{View synthesis.}
\end{abstract}

\begin{figure}
  \label{fig:figure0}
  \centering
  \includegraphics[width=0.9\columnwidth,height=0.6\columnwidth]{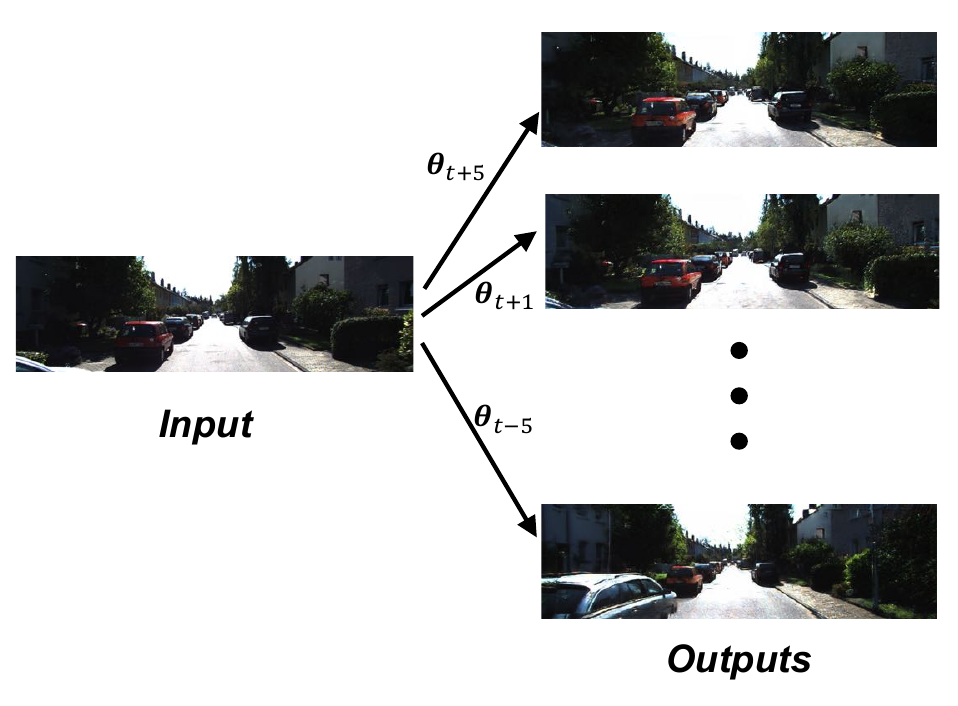}
  \caption{Illustration of our purpose. Our method aims to synthesize high-quality target images with the corresponding control variables $\bm{\theta}_{t+i}$. The output images are the synthesized results from our method.}
  \label{fig:figure0}
\end{figure}

For geometry-based methods, advance knowledge of the 3D geometry of the original image is required. It is difficult to collect the dense point clouds necessary for 3D construction. Moreover, structural estimation from a single image outdoors is also a  difficult research direction. The regions that are unseen in the input views due to self-occlusions must be deduced, otherwise, there will be holes in the generated image. Consequently, post-processing using an additional hole-filling algorithm is necessary \cite{Zhu2016Depth}. Therefore, this kind of algorithm may not always be feasible.

In learning-based approaches, the attributes of a certain objects are learned, such as symmetry in shapes. Thus, an image of this object from a novel view can be obtained based on prior knowledge. Methods of this kind are restricted by the availability of prior knowledge related to the relevant object category, such as cars or chairs. For large-scale environments, such as outdoor environments, the generated images from novel views may be distorted because these methods do not explicitly consider structural information. The reason for including structural information is that position and orientation of each object will incur different pixel allocations during egomotion of the camera. 

In this work, we combine the strengths of geometry-based and learning-based approaches. We introduce a new parametric image generation method for novel image synthesis based on corresponding camera transformation. The structural information of environments is explicitly considered in our framework. We incorporate the geometric information into the feature maps of our networks instead of constructing the 3D structure of the environments as an intermediate stage. The mean squared error is normally adopted as the loss function for view synthesis problem. Blurry image will be obtained because of the random filling of missing (unseen) regions. To fill in the missing parts and textures and improve the quality of the generated images, generative adversarial networks are applied as a loss function learning mechanism. The transformation constraint is treated as an auxiliary term in our adversarial losses.

The main contributions of our paper are summarized as follows.
\begin{enumerate}
    \item We propose a novel view synthesis method for large-scale scenes by incorporating inverse depth features. The inverse depth features are obtained by networks trained with sparse labeled depth values. Structural information is explicitly considered in our framework.
    \item We extend the proposed framework to fuse arbitrary number of input images with corresponding viewpoints.
    \item The proposed framework can be trained in an end-to-end way. Images of arbitrary size can be used as inputs because our networks are fully convolutional.
    \item We apply GAN to improve the quality of generated images, and transformation constraint is used as auxiliary term in our discriminator.
    % \item Max selection in features maps to fuse  verified the interpretability of hidden nodes' representations in feature space.
\end{enumerate}

The structure of this paper is organized as follows. In section 2, we introduce the related works about view synthesis problem and generative adversarial networks. In section 3, we introduce our framework for view synthesis with single and multiple inputs. Experiments of novel view synthesis using single and multiple images are reported and analyzed in section 4. We conclude our paper in the last section.

\section{Related Works}

The ability to predict the future movements of a robots' surroundings is one of the crucial requirements for decision-making and planning in robotics. Most related works anticipate the movement of certain objects in the scene under the assumption that the background is fixed \cite{finn2016unsupervised}. However, in the real world, robot's view constantly changing as it moves, which introduces additional challenges when solving this problem.

Deep neural networks have achieved great success in recent years because of their representation power. Research on novel view synthesis using convolutional neural networks can be categorized in two main directions.
%In order to generate the next frame image, the difference between two consecutive frame is considered. Optical flow between two frame is normally applied for the image generation. Some work try to learn the relationship between the two frames directly, Long short term memory (LSTM) is widely used for

\textbf{Learning depth values for view synthesis} For geometry-based view estimation method, reprojection is the most commonly applied for image synthesis. The quality of the generated image depends on the precision of estimated depth values. To generate entire image, a dense depth map is required. Currently, inferring depth values from images captured outside is still a challenging problem \cite{Eigen2014Depth,liu2016learning}. Furthermore, this kind of method cannot infer information about the regions that are missing in the input images. Consequently, holes will appear in the output image \cite{Mahjourian2017Geometry}. The geometry inferred  from a single image is uncertain, and as a result, unrealistic and jarring rendering artifacts will occur in the generated view.

\textbf{Disentangling pose features} Auto-encoders are mainly applied to incorporate control variables \cite{Tatarchenko16Multi}. Variables that control the view changes associated with hidden nodes of objects are decoded for image generation. In  previous researches, objects such as cars and chairs have been applied for view synthesis. Gated Boltzmann machines have also been proposed to predict the transformation controlled by the multiplication of certain latent variables \cite{Memisevic2007Unsupervised}. Synthetic images of chairs can be generated from given poses \cite{Dosovitskiy2015Learning}, where the pose parameters are treated as the attribute of the model. However, framework of this kind cannot be applied for large-scale image generation, because it does not consider the structural information. Depth regression as structural constraints does not help. 

\textbf{Image rendering} The goal of image-based rendering is to synthesize a new view of scene via warping from nearby input images. Learning and generation of selection mask in image space is an effective strategy. Interpolation between views is achieved by learning the weight or mask of pixels in different views \cite{Flynn2016Deep, ji2017deep}. The appearance flow method is based on calculating the allocation of the pixels in the output view \cite{Zhou2016View, Park2017transformation}. This kind of methods cannot infer the image information from regions that are missing in the original images. In addition, the structure of the generated images cannot be preserved, meaning that distortion will inevitably occur.

\textbf{Generative adversarial networks} Generative adversarial networks (GAN) are generative models via an adversarial process, which have achieved impressive performances in various tasks \cite{Goodfellow2014Generative, salimans2016improved}. GAN contains two models: generator G and discriminator D. It corresponds to a minimax two-player game. The generator tries to capture the data distribution and generate realistic data. On the other hand, the discriminator ties to distinguish whether data is generated or real. It is believed the GAN framework can be used to learn the loss function via the training process \cite{pix2pix2017}. Many methods for improving the quality of generated images have been proposed, such as Least Squares GAN, WGAN \cite{arjovsky2017wasserstein} and etc. The conditional GAN approach has been proposed to obtain images based on conditional attributes or images \cite{Mirza2014Conditional}. Based on this structure, several applications and modifications are introduced, such as image translation framework \cite{pix2pix2017}. Auxiliary classifier GAN (AC-GAN) \cite{Augustus2017AC} uses the conditional variables by designing additional classifier in discriminator.

%-------------------------------------------------------------------------
\section{Method}

In this section, we present our framework for novel view synthesis. Our goal is to synthesize high-quality images from novel views based on single or multiple input images and specified transformation variables. For the generation of images of large-scale scenes, we propose a novel framework for calculating the location of features by incorporating inverse depth features as geometric information. We adopt this approach because according to our daily observations, the displacements of pixels are related to the inverse depths of corresponding objects. In comparison with far objects, the displacements of pixels representing near objects are larger related to the egomotion of camera. This is well-studied and described as instantaneous motion model in image space.

\subsection{Camera motion and feature allocation}

The instantaneous motion model describes the optical flow field as a function of the egomotion of camera and inverse depth value \cite{Heeger1992Subspace, Strecha2002Motion}. The vector describing the camera's egomotion, $\bm{\theta}_t=(\mathbf{\Omega};\mathbf{t})$, consists of rotation and translation vectors. The rotation vector is denoted by $\mathbf{\Omega}=\left(\Omega_x, \Omega_y, \Omega_z \right)^T$, and the translation vector is denoted by $\mathbf{t}=\left(t_x, t_y, t_z \right)^T$. The displacement of a pixel $u$ can be expressed as a function of its inverse depth value $d$ and motion parameters.

\begin{equation*}
  \mathbf{u}=\mathbf{Q}_{\Omega}\mathbf{\Omega}+d\mathbf{Q}_t \mathbf{t}
\end{equation*}
where 
\begin{equation*}
  \mathbf{Q}_\Omega=\left(\begin{array}{ccc}
  \frac{\tilde{x}\tilde{y}}{f} & -f-\frac{\tilde{x}^2}{f} &  \tilde{y}\\
  f+\frac{{\tilde{y}}^2}{f} & -\frac{\tilde{x}\tilde{y}}{y} & -\tilde{x} \\
  \end{array}\right)
\end{equation*}

\begin{equation*}
  \mathbf{Q}_t=\left(\begin{array}{ccc}
  -f & 0  &  \tilde{x}\\
  0 & -f & \tilde{y} \\
  \end{array}\right)
\end{equation*}
Here, $\tilde{x}=x-x_0$ and $\tilde{y}=y-y_0$ are the centered image coordinates, $f$ denotes the focal length, and $(x_0, y_0)$ represents the coordinates of the principal point. The pixel skew and aspect ratio in the camera calibration matrix is set to $0$ and $1$, respectively.

The displacement of a pixel consists of two components. The first component depends on the rotation of the camera and the second component depends on the inverse depth values and the translation of the viewpoint. This description is consistent with our observation that for an object at a large distance, the translation of the camera will result in a small displacement, and vice versa.
\begin{figure}
  \centering
  \includegraphics[width=0.9\columnwidth,height=0.5\columnwidth]{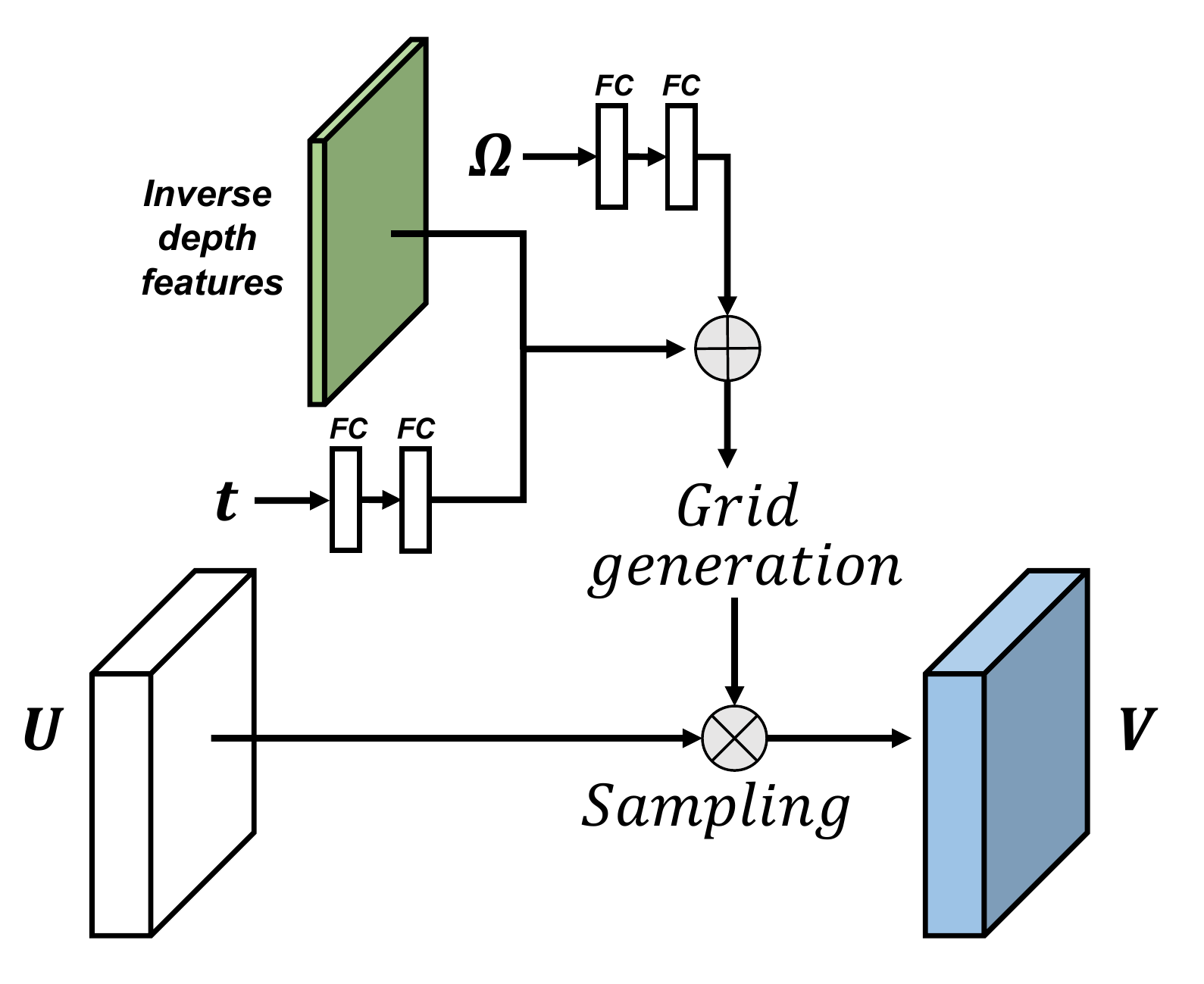}
  \caption{The basic structure of sampling grid generation framework. The sampling grid is generated using inverse depth features and control variables (Euler angles and translations along the x, y and z axes). $\mathbf{U}$ and $\mathbf{V}$ denote the input and output feature maps, respectively.}
  \label{fig:figure1}
\end{figure}

We sample the output features $\mathbf{V} \in \mathbb{R}^{H\times W \times C}$ from the input feature map $\mathbf{U}\in \mathbb{R}^{H\times W \times C}$ using parameterized sampling grid $\mathbf{G}\in \mathbb{R}^{HW \times 2}, G_i=(x_i^s, y_i^s)^T$, where $H$ and $W$ are the height and width of the grid, respectively, and C is the number of channels. Each node in the output feature map is obtained through this process.

The pointwise transformation is defined as follows:
\begin{equation}
\label{eq:grid_gen}
  \left(\begin{array}{c}
  x_i^s \\
  y_i^s \\
\end{array}\right)=g(\mathbf{\Omega})\left(\begin{array}{c}
    {x_i^t} \\
    {y_i^t}  \\
    {x_i^t}^2 \\
    {x_iy_i}  \\
    {y_i^t}^2 \\
    1 \\
  \end{array}\right)+d(x_i^s,y_i^s) h(\mathbf{t})\left(\begin{array}{c}
    x_i^t \\
    y_i^t \\
    1 \\
    \end{array}\right)
\end{equation}
where $(x_i^t,y_i^t)$ and $(x_i^s,y_i^s)$ denote the target and source coordinates in the output and input feature maps, respectively, and $d(x_i^s,y_i^s)\in[0,1]$ denotes the inverse depth feature corresponding to $(x_i^s,y_i^s)$ in the feature space. $d(x_i^s,y_i^s)$ is initialized with $d(x_i^t,y_i^t)$, and updated to $d(x_i^s,y_i^s)$ after one iteration of Equation \ref{eq:grid_gen}. The coordinates $x_i^s$, $y_i^s$, $x_i^t$, and $y_i^t$ are normalized to lie in the range $[-1, 1]$. $g(\mathbf{\Omega})\in \mathbb{R}^{2\times6}$ is a parametric transformation in terms of the rotation variables $\mathbf{\Omega}$. $h(\mathbf{t})\in \mathbb{R}^{2\times3}$ is a parametric transformation in terms of the translation variables $\mathbf{t}$. The transformation matrices are generated via fully connected networks. The basic structure of the sampling grid generation framework is shown in Figure \ref{fig:figure1}.

We apply the bilinear sampling kernel for feature sampling from the input feature map \cite{Jaderberg2015Spatial}, which is defined as:
\begin{equation*}
  V_i^c=\sum^H_h\sum^W_w U_{hw}^c \max{\left(0,1-|x_i^s-h|\right)}\max(0,1-|y_i^s-w|)
\end{equation*}

The gradients flow back to the input feature map and the sampling grid by virtue of the differentiable sampling mechanism.

\begin{figure*}
  \centering
  \includegraphics[width=0.9\columnwidth,height=0.5\columnwidth]{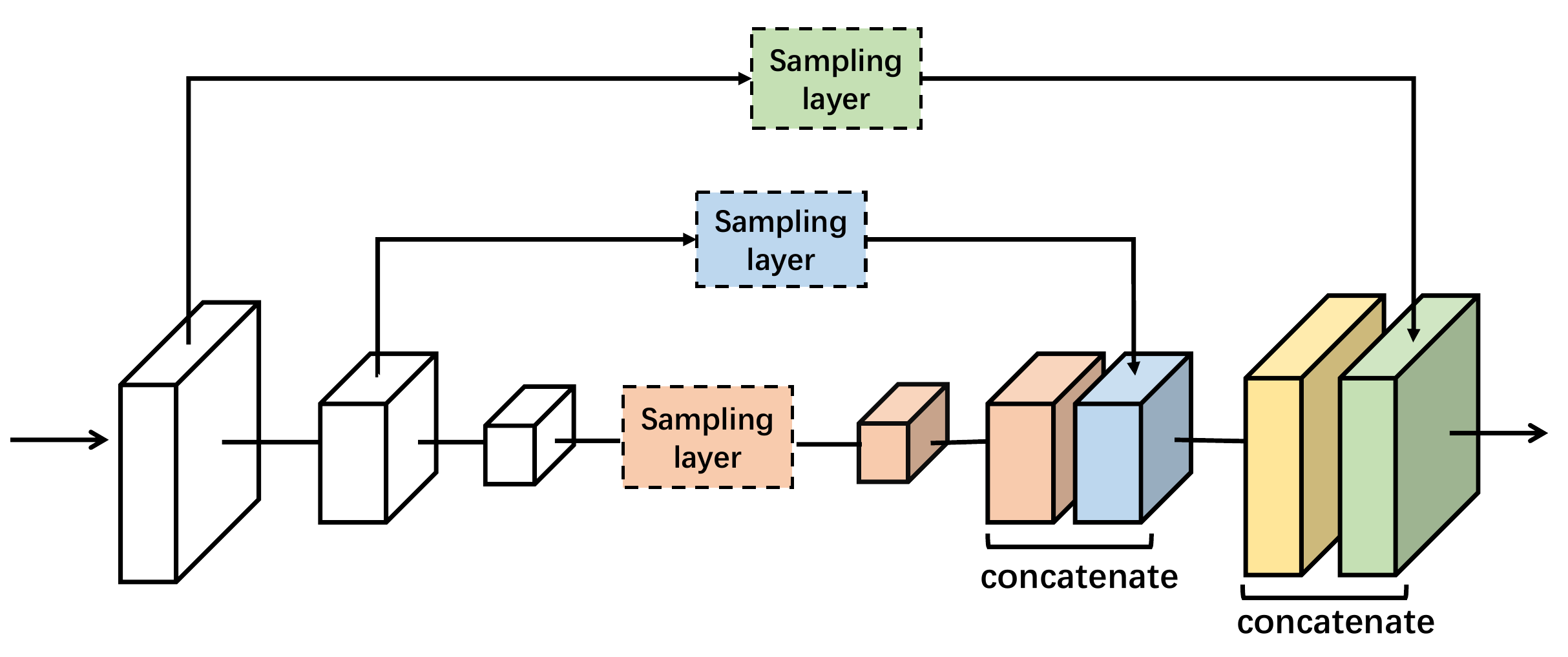}
  \caption{Structure of our image generation network. We concatenate the features at different representation levels.}
  \label{fig:figure2}
\end{figure*}

Our network structure is shown in Figure \ref{fig:figure2}. We stack transformations at different representation levels. The networks can be trained with $L_1$ loss, supervised by the corresponding output image. The generated images with $L_1$ loss function may incur blur for uncertain parts. Because view synthesis trained with the $L_1$ loss cannot infer the unseen parts and textures, the image generated in this way may suffer from blur in certain regions. Hence, generative adversarial networks are introduced to fill in the textures and unseen parts in the generated images.

% We follow the work of \cite{zhou2017unsupervised} and introduce the mask $\mathbf{M}$ to generate weight to eliminate loss gradients from unseen object. In our situation, we applied the depth features to provide the structural information. The penalty term to ensure the smoothness of depth is not involved in our algorithm.

% The mask term is generated from the objective function of L1 loss is define as

% \begin{equation}
%   \mathcal{L}_{L1} = \mathbb{E} \left[ M \odot \lVert Y-G(X,\bm{\theta},D_{1,2,3},z)\rVert_{1}\right]
% \end{equation}.

% In order to generate the desired pictures, we also need to consider the structural information of environment. From our observation, the position of object in the next frame depends on the distance between the camera. The near object move more drastically than the ones' far away from the camera.

% The mask term is generated from the warping coordinates with small networks. The mask is the output of softmax layer with label 1. In order to regularize the output of mask, cross-entropy is applied as objective function.

\subsection{Generative adversarial networks}

The $L_1$ loss function captures the main structure of the target images. Unseen regions and textures will cause blurring because all the possibilities for the unknown pixels will be mixed. To synthesize realistic images, generative adversarial networks are introduced to learn the loss function to fill in missing textures and occluded objects. Furthermore, in addition to improving the quality of images, auxiliary loss is introduced to ensure that the generated images satisfy the specified transformation constraint. Based on these requirements, the Auxiliary Classifier GANs (AC-GAN) framework is applied to construct our discriminator. Least squares loss function (LSGAN) \cite{Mao2017lsgan} is used as part of our AC-GAN to learn the image distribution, which is used to judge whether an input image is real or not. Compared with the regular GAN, LSGAN can generate more realistic images.

% \begin{figure}[htbp]
%   \centering
%   \includegraphics[width=0.5\columnwidth,height=0.6\columnwidth]{figure3.pdf}
%   \caption{Structure of our generator networks. Image is generated from a single input image and the control variable vector $\bm{\theta}_t$. The inverse depth features of input image are provided in the branch networks.}
%   \label{fig:figure3}
% \end{figure}
% \begin{figure}[htbp]
%   \centering
%   \includegraphics[width=0.4\columnwidth]{figure4.pdf}
%   \caption{Adversarial loss with a real input image classifier and the real pose $\bm{\theta}_t$ regression.}
%   \label{fig:figure4}
% \end{figure}

% \begin{figure}[htbp]
%   \centering
%   \includegraphics[width=0.5\columnwidth]{figure5.pdf}
%   \caption{Adversarial loss with a generated image classifier and the fake random pose variables $\mathbf{z}_p$ regression.}
%   \label{fig:figure5}
% \end{figure}

\begin{figure}[htbp]
\centering
\subfigure[Generator]{\includegraphics[width=0.26\columnwidth,height=0.25\columnwidth]{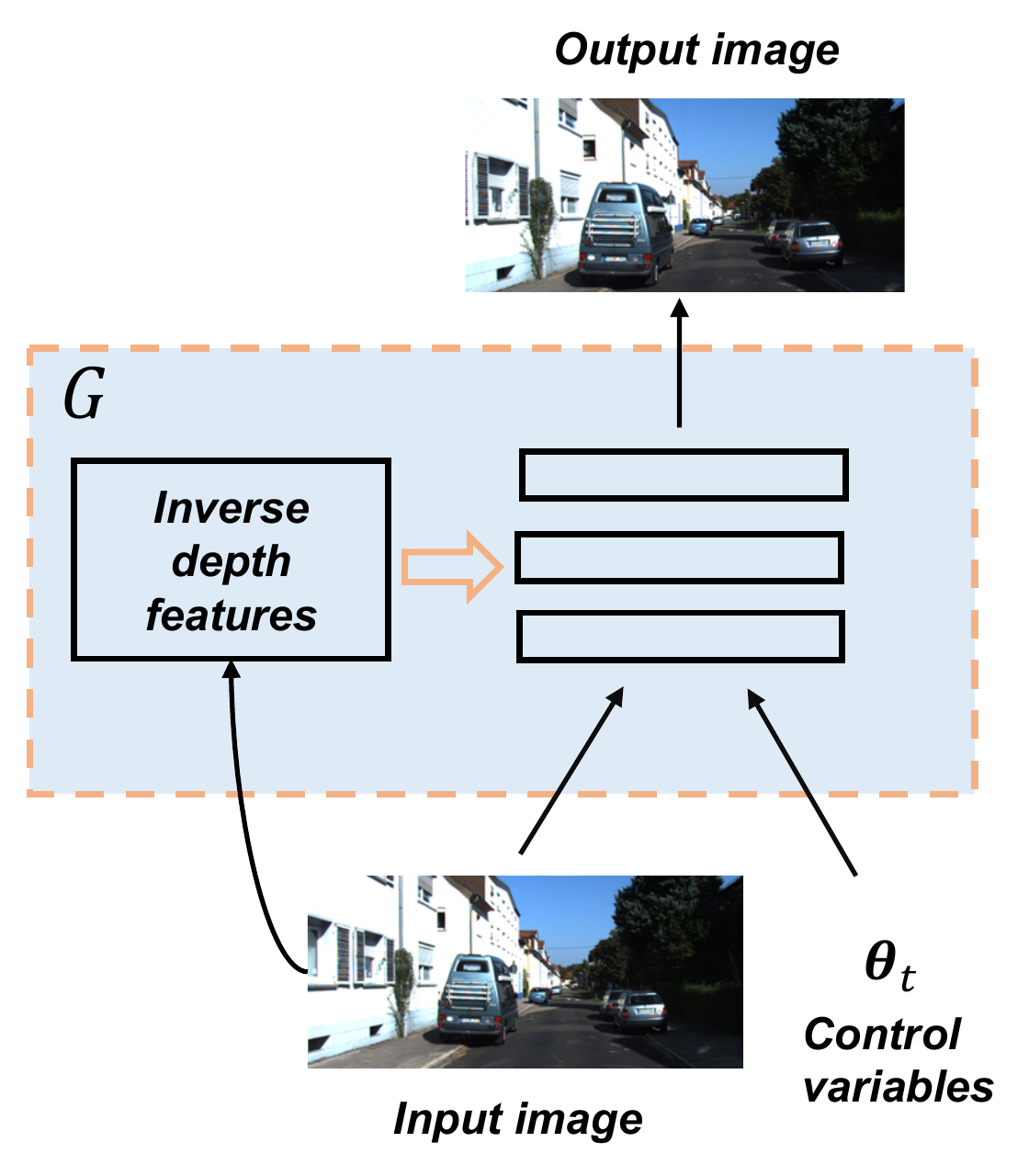}}                                                          
\subfigure[Real inputs]{\includegraphics[width=0.312\columnwidth,height=0.18\columnwidth]{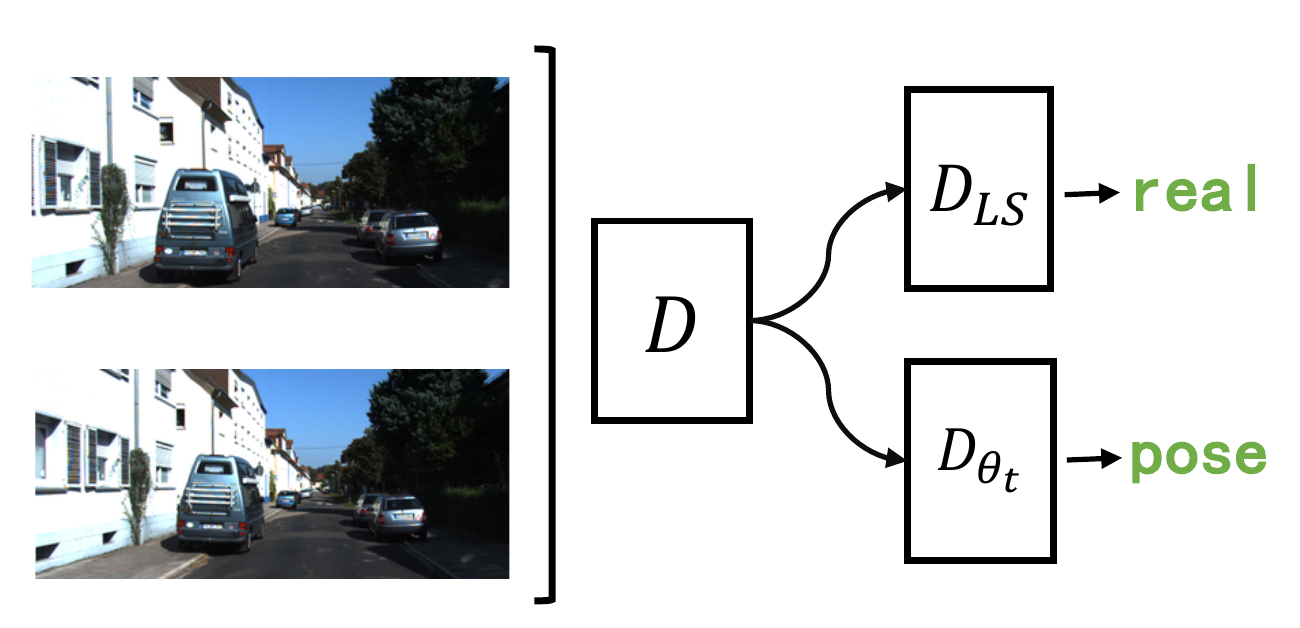}}
\subfigure[Fake inputs]{\includegraphics[width=0.412\columnwidth,height=0.18\columnwidth]{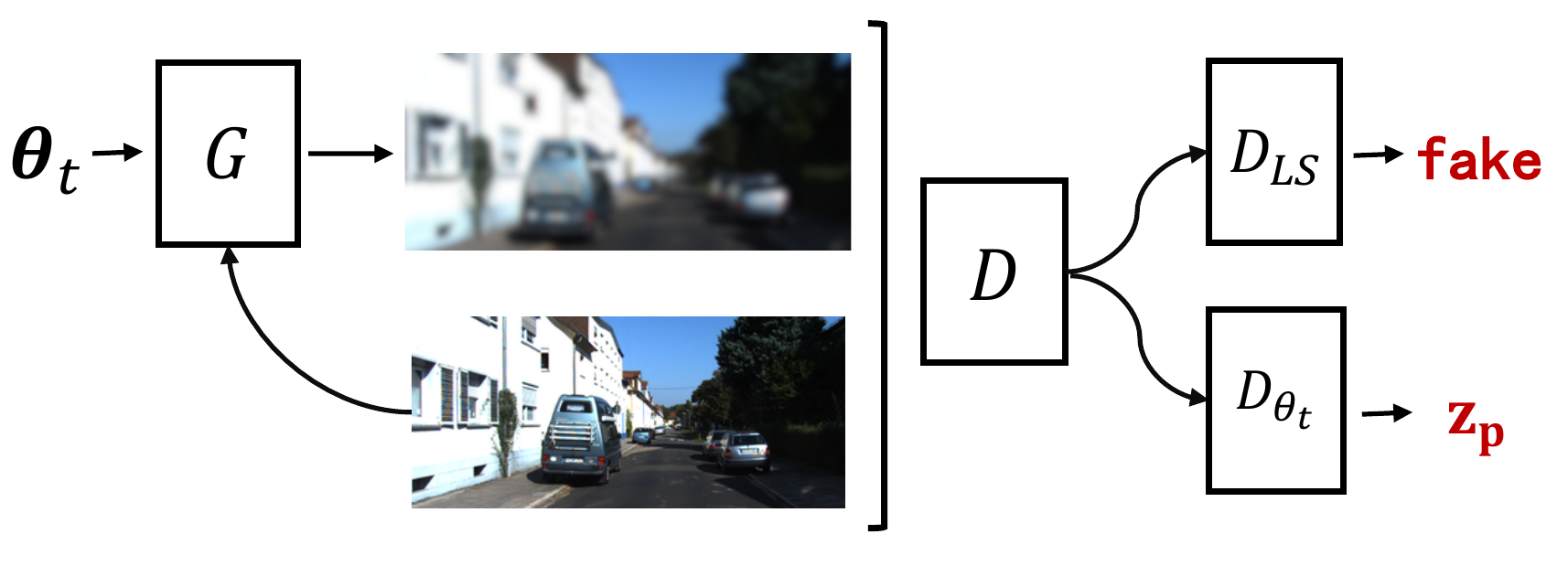}}
\caption{Structure of our generator networks. Image is generated from a single input image and the control variable vector $\bm{\theta}_t$. The inverse depth features of input image are provided in the branch networks. Adversarial loss with a real input image classifier and the real pose $\bm{\theta}_t$ regression. Adversarial loss with a generated image classifier and the fake random pose variables $\mathbf{z}_p$ regression.}
\end{figure}

The generator $G$ and discriminator $D$ are obtained by minimizing the objective function of generative adversarial networks. The objective function of discriminator consists of two terms: least squares term and auxiliary term. The least squares term models the distribution of image, and it determines whether the generated images are real or fake. The objective function of the least squares term is defined as follows:

\begin{equation}
  \begin{aligned}
    \mathcal{L}_{LS}=&\frac{1}{2} \mathbb{E}_{Y\sim p_{data(Y)}} \left[ \left(D_{LS}(Y|X)-1\right)^2\right]+\\
    &\frac{1}{2}\mathbb{E}_{z\sim p_{z(z)}} \left[ \left(D_{LS}(G(X,\bm{\theta_t},z)|X)+1\right)^2\right]
  \end{aligned}
\end{equation}
where $z\sim p_{z(z)}$ denotes random values, which bring randomness to fill the unseen regions and textures. In our work, dropout layers are applied to introduce randomness, it is employed after the sampling layer in training and test phases.

The auxiliary term works as regularization term. It is applied to ensure the two input images satisfy the specific requirements of transformation. The objective function of auxiliary term is expressed below:
\begin{equation}
  \begin{aligned}
    \mathcal{L}_{\bm{\theta}_t}=& \frac{1}{2} \mathbb{E}_{Y\sim p_{data(Y)}}\left[(D_{\theta_t}(Y|X)-\bm{\theta_t})^2\right]+\\
     &\frac{1}{2}\mathbb{E}_{\bf{z}_t\sim \mathcal{N}(0,1)}\left[(D_{\theta_t}(G(X,\bm{\theta_t},z)|X)-\bf{z}_t)^2\right]
  \end{aligned}
\end{equation}
%    \mathbb{E}}_{\bf{z}_t\sim \mathcal{N}(0,1)\left[(D_{\theta_t}(G(X,z)|X)-\bf{z}_t)^2\right]
The final objective function of discriminator is:
\begin{equation}
    \min_{D_{LS},D_{\bm{\theta}_t}}\mathcal{L}_{AC-GAN}=\mathcal{L}_{LS}+ \lambda \mathcal{L}_{\bm{\theta}_t}
\end{equation}
where $\lambda$ is the weight of auxiliary term, and $\lambda$ is set to $0.1$.

The final objective function of generator is defined as follows:
\begin{equation}
  \begin{aligned}
    \min_G \mathcal{L}_{AC-GAN} = &\frac{1}{2} \mathbb{E}_{z} \left[ (D_{LS}(G(X,\bm{\theta_t},z)|X)^2  \right]+\\
    &\lambda \mathbb{E}\left[(D_{\theta_t}(G(X,\bm{\theta_t},z)|X)-\bm{\theta_t})^2\right] + \\
    &\varphi \mathbb{E}\left[|G(X,\bm{\theta_t},z)-Y|\right]
  \end{aligned}
\end{equation}
where $\varphi$ is the weight of $L_1$ loss term. 

The structure of generator networks is shown in Figure \ref{fig:figure3}. A single image and a vector of control variables are taken as the inputs. The discriminator's structure of our framework is illustrated in Figure \ref{fig:figure4} and Figure \ref{fig:figure5}. Both the generated and original input images are taken as the inputs for discrimination. For our task, in addition to generate high-quality images, the specified transformation should also be satisfied. Hence, there are two branches  for our discriminator's network. The purpose of one branch is to ensure the generation of realistic image, the purpose of the other branch is to ensure the generated image meets the transformation requirements. For a synthesized input image, random values $\mathbf{z}_p$ following a normal distribution are used as the supervised signals. 

\subsection{Fusion of input views}

Image from a single view may not contain adequate information for view inference. Information obtained from different views can provide supplementary details for novel view synthesis. Therefore, we extend our method as described above to the fusion of multiple input images from different viewpoints.

Max selection is introduced in our framework to fuse features from multiple feature maps. The objective function for max selection is expressed as follows:

\begin{equation}
  % V_{out}(\mathbf{p})=\max\left (V_1(\mathbf{p}),V_2(\mathbf{p}),\ldots,V_N(\mathbf{p})\right), \forall p\in S
  V_{out}(p)=\max\left (V_1(p),V_2(p),\ldots,V_N(p)\right), \forall p\in S
\end{equation}
where $V$ is the feature map after performing grid sampling, $V:S\subset \mathbb{R}^3 \mapsto \mathbb{R}$, $p=(x,y,c)$ denotes spatial location $(x,y)$ in channel $c$, and $N$ is the number of feature maps to be fused.

\begin{figure}[htbp]
  \centering
  \includegraphics[width=\columnwidth,height=0.6\columnwidth]{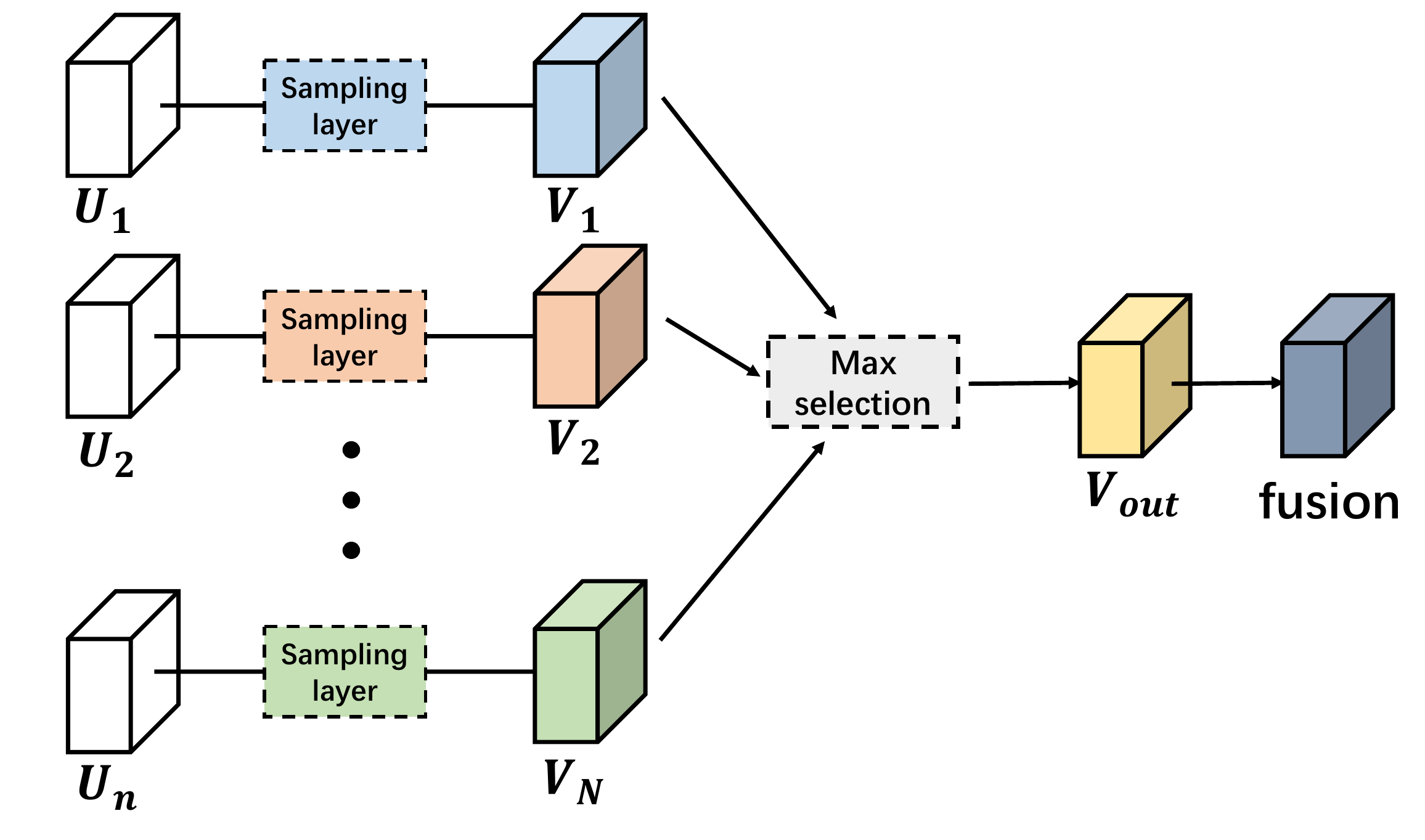}
  \caption{Feature fusion network for the fusion of multiple images from different corresponding viewpoints.}
  \label{fig:figure6}
\end{figure}

The network structure for image fusion is shown in Figure \ref{fig:figure6}. Our method does not restrict the number of inputs. We apply the max selection after the sampling layer. The maximum element at each position is selected as the output.

\subsection{Network architecture}

We employ DenseNet-201 as the backbone model without pooling layers \cite{Huang2017Densely}. This architecture connects each layer to every other layer in a feed-forward fashion. By virtue of this feature, DenseNets alleviate the vanishing-gradient problem, strengthen feature propagation and reuse features of lower layers, and substantially reduce the number of parameters. 

We sample and transform the features in transition layers of DenseNet. In order to include more details of input images to the synthesized images, features of different levels in transition layers are sampled and concatenated. Otherwise, words on traffic sign and some small objects in the input images will not appear on the synthesized images. Both inverse depth estimation networks and image generation networks are based on DenseNet-201. Parameters of inverse depth estimation networks are fixed during the training and testing phases of image synthesis networks, it is used to provide the inverse depth feature maps at different levels. 

\section{Experiments}

In this section, we conduct experiments to verify the effectiveness of our method on the KITTI dataset \cite{Geiger2012CVPR}, which is collected in urban environments. The KITTI dataset consists of image sequences from a driving vehicle with depth captured by a LiDAR and ground truth position by a GPS localization system. The depth values obtained from LiDAR are sparse, and only 5\% of pixels contain valid values. Our method can predict the picture of novel view up to move $\pm7$ meters along the z axis and rotate around the y axis by $\pm22$ degrees.

Our neural networks are trained and deployed on a NVIDIA TITAN X GPU with 12 GB memory. We use PyTorch to train our model. Our network is modified based on  DenseNet-201 downloaded from model zoo. We apply the images in sequence 00 and 02 as training set (91960 image pairs). Images in sequence 08, 09, 10 are taken as test set (68540 image pairs). Images of different sequences are collected in different urban environments. Therefore, environments in the test set will not appear in the training set. All the image are down-sampled by a factor of 2 and resized to 176x608. The Euler angles and translations along the x,y and z axes are applied as inputs to our networks.

The inverse depth values are preprocessed by normalizing the values to $\frac{2}{x-1.5}-1$, where $x$ is the depth value collected in LiDAR. All layers are trained using SGD and ADAM optimizer \cite{Kingma2014Adam} ($\beta_1=0.5, \beta_2=0.9$), and learning rate is set to 0.0001. Our networks are trained in an end-to-end way after 200,000 iterations.

In order to evaluate the effectiveness of our method, we designed the experiments for novel view synthesis for large-scale scene. We testify our framework for generating novel view image from a single input image, multiple input images. Our method is evaluated quantitatively using mean pixel $L_1$ error and inception score \footnote{The code and model for inception score evaluation are available at \url{https://github.com/openai/improved-gan}}. These two metrics are widely applied to evaluate the synthesized images.

\subsection{View synthesis from a single input}

\begin{figure*}[htbp]
\centering
\setcounter{subfigure}{0}
\subfigure[Ground truth]{\label{fig:gt_single}
\begin{minipage}[t]{0.175\textwidth}
\includegraphics[width=1.0\textwidth,height=0.35\columnwidth]{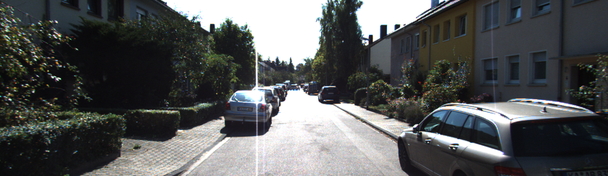}\vspace{0.5ex}
\includegraphics[width=1.0\textwidth,height=0.35\columnwidth]{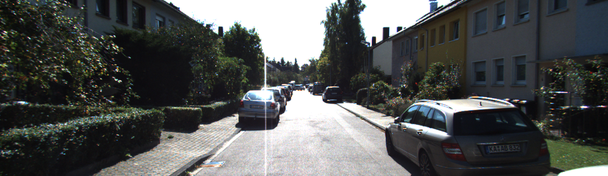}\vspace{0.5ex}
\includegraphics[width=1.0\textwidth,height=0.35\columnwidth]{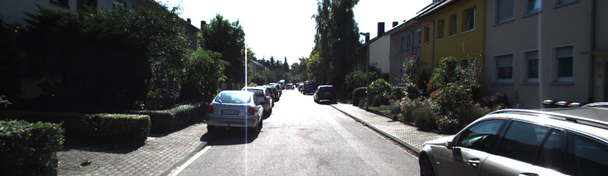}\vspace{0.5ex}
\includegraphics[width=1.0\textwidth,height=0.35\columnwidth]{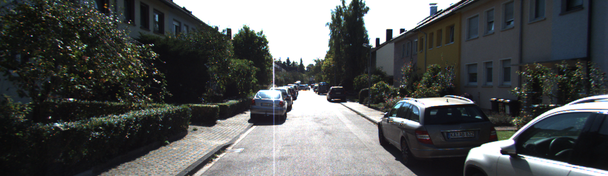}\vspace{0.5ex}
\includegraphics[width=1.0\textwidth,height=0.35\columnwidth]{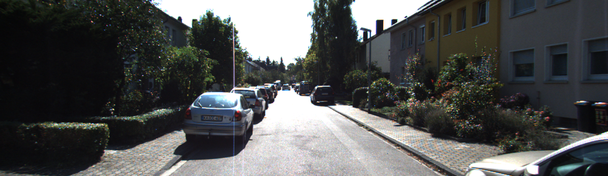}\vspace{0.5ex}
\includegraphics[width=1.0\textwidth,height=0.35\columnwidth]{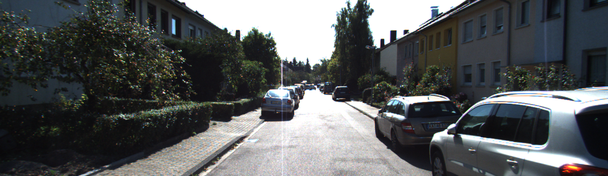}\vspace{0.5ex}
\end{minipage}
}
\subfigure[MV3D]{\label{fig:mv3d_single}
\begin{minipage}[t]{0.175\textwidth}
\includegraphics[width=1.0\textwidth,height=0.35\columnwidth]{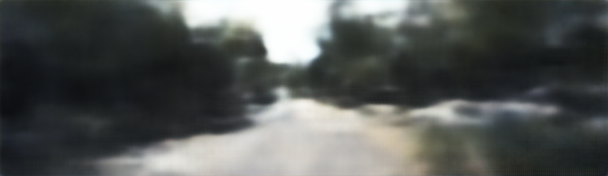}\vspace{0.5ex}
\includegraphics[width=1.0\textwidth,height=0.35\columnwidth]{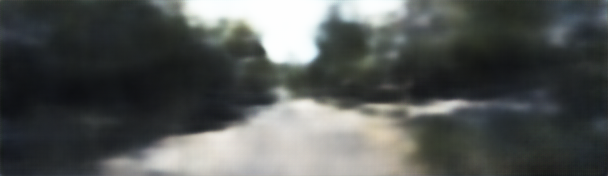}\vspace{0.5ex}
\includegraphics[width=1.0\textwidth,height=0.35\columnwidth]{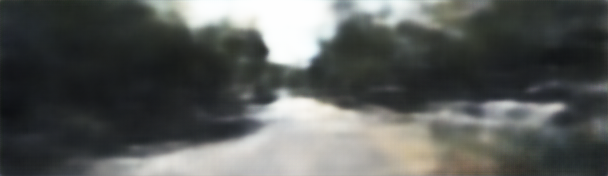}\vspace{0.5ex}
\includegraphics[width=1.0\textwidth,height=0.35\columnwidth]{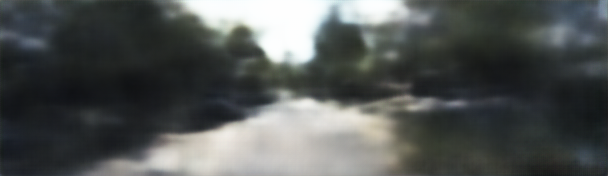}\vspace{0.5ex}
\includegraphics[width=1.0\textwidth,height=0.35\columnwidth]{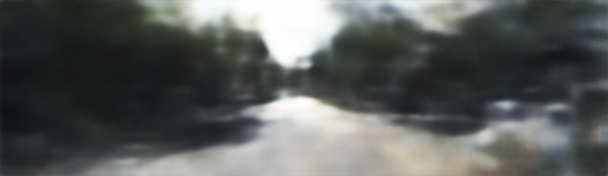}\vspace{0.5ex}
\includegraphics[width=1.0\textwidth,height=0.35\columnwidth]{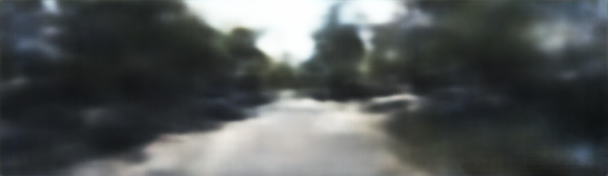}\vspace{0.5ex}
\end{minipage}
}
\subfigure[Appearance flow]{\label{fig:af_single}
\begin{minipage}[t]{0.175\textwidth}
\includegraphics[width=1.0\textwidth,height=0.35\columnwidth]{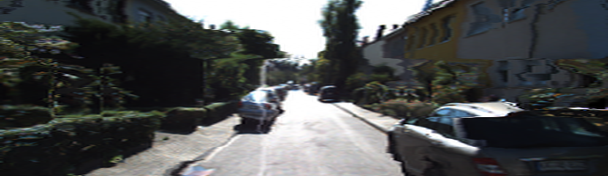}\vspace{0.5ex}
\includegraphics[width=1.0\textwidth,height=0.35\columnwidth]{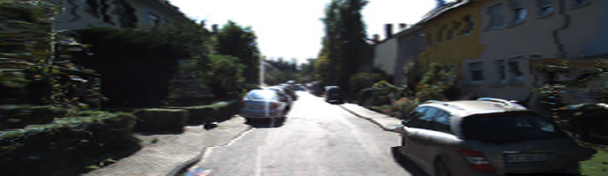}\vspace{0.5ex}
\includegraphics[width=1.0\textwidth,height=0.35\columnwidth]{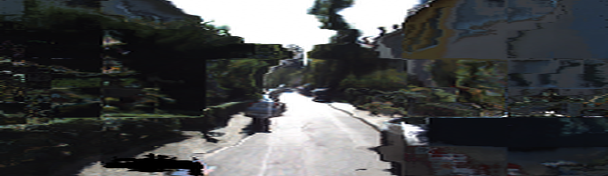}\vspace{0.5ex}
\includegraphics[width=1.0\textwidth,height=0.35\columnwidth]{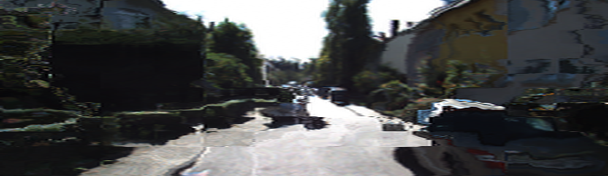}\vspace{0.5ex}
\includegraphics[width=1.0\textwidth,height=0.35\columnwidth]{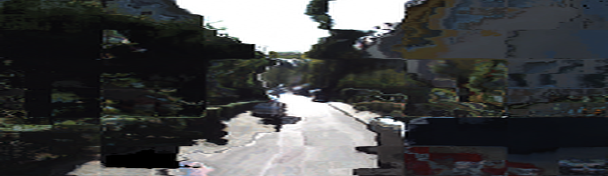}\vspace{0.5ex}
\includegraphics[width=1.0\textwidth,height=0.35\columnwidth]{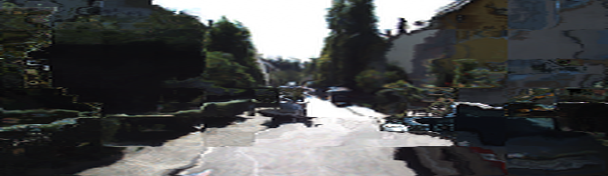}\vspace{0.5ex}
\end{minipage}
}
\subfigure[Our networks ($L_1$)]{\label{fig:l1_single}
\begin{minipage}[t]{0.175\textwidth}
\includegraphics[width=1.0\textwidth,height=0.35\columnwidth]{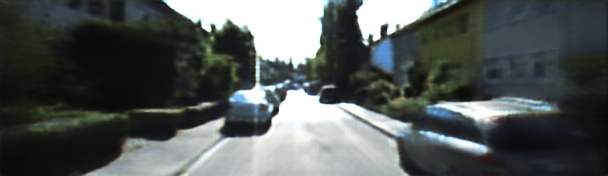}\vspace{0.5ex}
\includegraphics[width=1.0\textwidth,height=0.35\columnwidth]{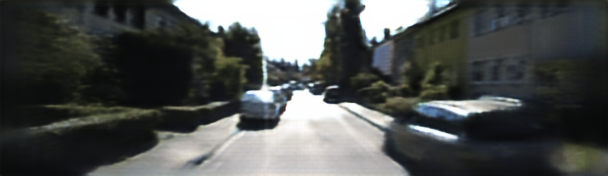}\vspace{0.5ex}
\includegraphics[width=1.0\textwidth,height=0.35\columnwidth]{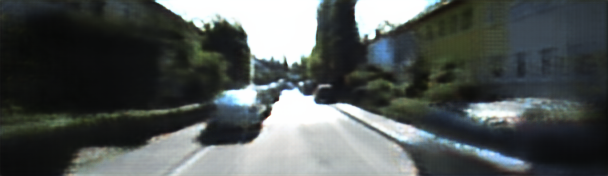}\vspace{0.5ex}
\includegraphics[width=1.0\textwidth,height=0.35\columnwidth]{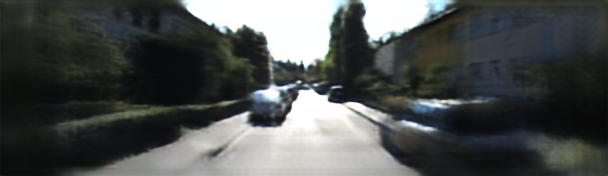}\vspace{0.5ex}
\includegraphics[width=1.0\textwidth,height=0.35\columnwidth]{l1_plus_three_step_000525.png}\vspace{0.5ex}
\includegraphics[width=1.0\textwidth,height=0.35\columnwidth]{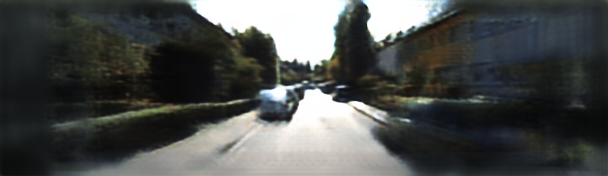}\vspace{0.5ex}
\end{minipage}
}
\subfigure[Our networks ($L_1$+GAN)]{\label{fig:gan_single}
\begin{minipage}[t]{0.175\textwidth}
\includegraphics[width=1.0\textwidth,height=0.35\columnwidth]{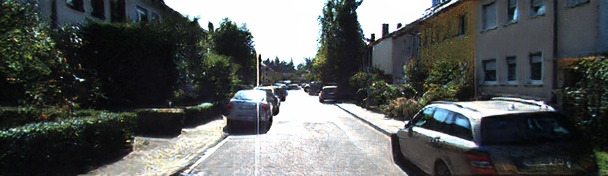}\vspace{0.5ex}
\includegraphics[width=1.0\textwidth,height=0.35\columnwidth]{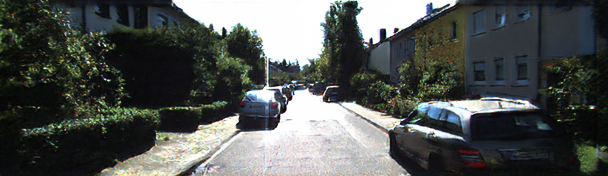}\vspace{0.5ex}
\includegraphics[width=1.0\textwidth,height=0.35\columnwidth]{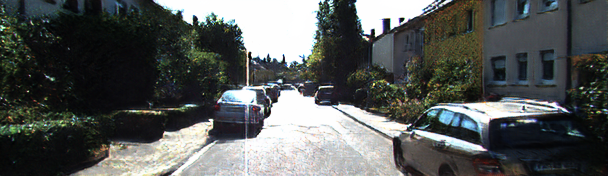}\vspace{0.5ex}
\includegraphics[width=1.0\textwidth,height=0.35\columnwidth]{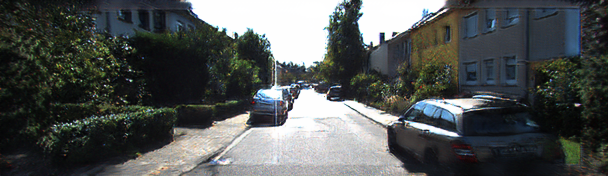}\vspace{0.5ex}
\includegraphics[width=1.0\textwidth,height=0.35\columnwidth]{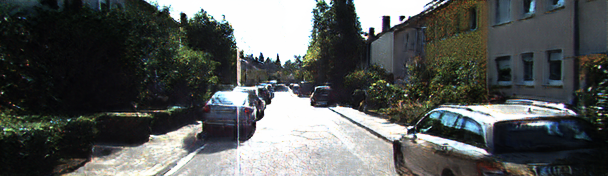}\vspace{0.5ex}
\includegraphics[width=1.0\textwidth,height=0.35\columnwidth]{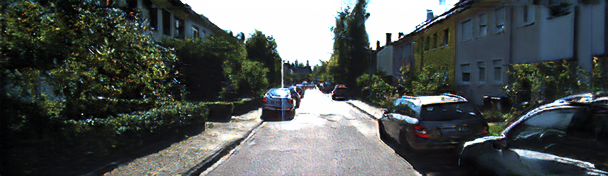}\vspace{0.5ex}
\end{minipage}
}
\caption{Examples of novel view synthesis results with single input image from the KITTI dataset. For each frame, we show (a) ground truth images, (b) results generated with MV3D method, (c) results generated with appearance flow method (d) results produced by our networks trained with $L_1$ loss, and (e) synthesized images by our networks with $L_1$ and GAN loss. Different rows shows the results of different time intervals. Images from the first row to the bottom row are one time step forward, one time step backward, three time steps forward, three time steps backward, five time steps forward and five time steps backward.}
\label{fig:view_synthesis_single_image}
\end{figure*}

Our method can predict the image from a single input image with the given viewpoint. We generate the results from ten adjacent time steps (five time steps forward and five time steps backward). $+$ and $-$ denote forward and backward movement respectively. Generated images of $\pm$ one time step, $\pm$ three time steps and $\pm$ five time steps are shown in Figure \ref{fig:view_synthesis_single_image}.

We evaluate our method with two parameterized methods: Multi-view 3D Models (MV3D) \cite{Tatarchenko16Multi} \footnote{The code and model are available at \url{https://github.com/lmb-freiburg/mv3d}} and appearance flow method \cite{Zhou2016View} \footnote{The code and model are available at \url{https://github.com/tinghuiz/appearance-flow}}. We also train our networks with $L_1$ Loss function. In order to maintain the structural information explicitly, MV3D preserves the structure of environments by adding the depth regression as additional output channel. However, this method is not effective for large-scale scenes, even for image synthesis with slight egomotion. The appearance flow method can keep the structural information and generate high-quality images for slight movements. However, images will become distorted for large movement. The generated images using MV3D and appearance flow methods on the KITTI dataset are consistent with the results demonstrated in \cite{Zhou2016View}. Figure \ref{fig:l1_single} shows the generated images with $L_1$ loss function, which demonstrate that our framework is an effective method to incorporate the structural information. For missing parts in the original images, it will become blurry in the corresponding parts in the output regions. Figure \ref{fig:gan_single} demonstrates that GAN loss can improve the quality of the generated images by filling the blurry parts with textures and details. Our method can also conjecture objects to fill the unseen parts.

Table \ref{tab:l1_single} shows the mean pixel $L_1$ errors of different time intervals of three methods. $L_1$ error measures the low frequency signals of images. Therefore, lower $L_1$ error means the method can capture more structural information of target images. Our framework trained with $L_1$ loss has achieved the lowest $L_1$ error. The final objective function of our method reaches a compromise between maintaining structural information and filling uncertain details by involving GAN loss and $L_1$ loss. By doing this, $L_1$ error of our generated images will increase. For all the methods, $L_1$ errors will increase as the time interval increase. The inception score of our method does not change.

\begin{table*}
\begin{center}
\begin{tabular}{l c c c c c c}
\toprule
\multirow{2}*{Method} & $\pm$ One  & $\pm$ Two  & $\pm$ Three  & $\pm$ Four & $\pm$ Five & \multirow{2}*{Mean}\\
&time step & time steps &  time steps & time steps & time steps\\
\midrule
MV3D \cite{Tatarchenko16Multi} &  0.1933 &  0.2057 &  0.2224 & 0.2411 & 0.2593 & 0.2244\\
Appearance Flow \cite{Zhou2016View}  & 0.2232  & 0.2850 & 0.3234 & 0.3508 &  0.3720 & 0.3109\\
Our networks ($L_1$) & \textbf{0.1458} &  \textbf{0.17548} & \textbf{0.1991} & \textbf{0.2203} & \textbf{0.2394} & \textbf{0.1959}\\
Our networks ($L_1$+GAN)  & 0.2057 &  0.2360 &0.2630 &  0.2885 & 0.3056 & 0.2598\\
\bottomrule
\end{tabular}
\end{center}
\caption{Mean pixel $L_1$ error on the KITTI dataset for novel view synthesis.}
\label{tab:l1_single}
\end{table*}

Inception scores of different time intervals are shown in Table \ref{tab:InceptionScore_single}. Distortion of the generated images does not affect the inception score. Therefore, both mean pixel $L_1$ error and inception score need to be taken as evaluation metrics for view synthesis methods.
\begin{table*}
\scriptsize
\begin{center}
\begin{tabular}{l c c c c c c}
\toprule
\multirow{2}*{Method} & $\pm$ One  & $\pm$ Two  & $\pm$ Three & $\pm$ Four & $\pm$ Five  & \multirow{2}*{Mean}\\
& time step & time steps &  time steps & time steps & time steps\\
\midrule
MV3D \cite{Tatarchenko16Multi} & 1.705$\pm$0.018 & 1.700$\pm$0.019
 & 1.702$\pm$0.014 & 1.720$\pm$0.017
 & 1.716$\pm$0.024
 & 1.709\\
Appearance  & \multirow{2}*{\textbf{3.196$\pm$0.038}} & \multirow{2}*{3.084$\pm$0.023} & \multirow{2}*{2.951$\pm$0.034}& \multirow{2}*{2.855$\pm$0.048}& \multirow{2}*{2.768$\pm$0.035} & 2.971\\
Flow \cite{Zhou2016View} & &  &  &  & \\

Our networks & \multirow{2}*{2.819$\pm$0.034} & \multirow{2}*{2.836$\pm$0.021} & \multirow{2}*{2.874$\pm$0.026} & \multirow{2}*{2.872$\pm$0.040} & \multirow{2}*{2.860$\pm$0.036} & \multirow{2}*{2.852}\\
($L_1$) & &  &  &  & \\
Our networks & \multirow{2}*{\textbf{3.196$\pm$0.054}} & \multirow{2}*{\textbf{3.203$\pm$0.053}} & \multirow{2}*{\textbf{3.196$\pm$0.038}} & \multirow{2}*{\textbf{3.173$\pm$0.036}}  & \multirow{2}*{\textbf{3.196$\pm$ 0.040}}& \multirow{2}*{\textbf{3.193}}\\
($L_1$+GAN) & &  &  &  & \\
\bottomrule
\end{tabular}
\end{center}\caption{Mean and standard deviation (std) of inception score on the KITTI dataset for novel view synthesis. Higher is better. For the real images on the KITTI dataset, the mean inception score is $2.9656$.}
\label{tab:InceptionScore_single}
\end{table*}

\subsection{View synthesis from multiple inputs}
\begin{figure*}[htbp]
\centering
\setcounter{subfigure}{0}
\subfigure[Ground truth]{\label{fig:gt_multi}
\begin{minipage}[t]{0.22\textwidth}
\includegraphics[width=1.0\textwidth,height=0.35\columnwidth]{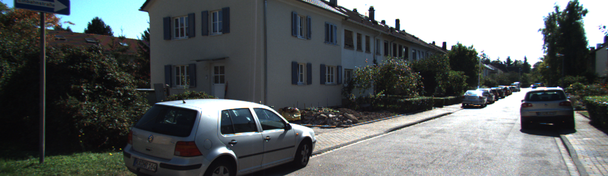}\vspace{0.5ex}
\includegraphics[width=1.0\textwidth,height=0.35\columnwidth]{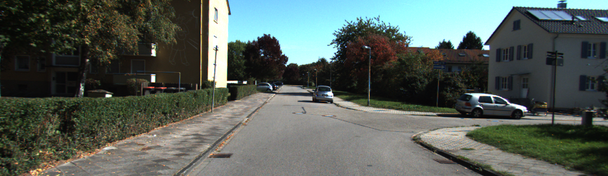}\vspace{0.5ex}
\includegraphics[width=1.0\textwidth,height=0.35\columnwidth]{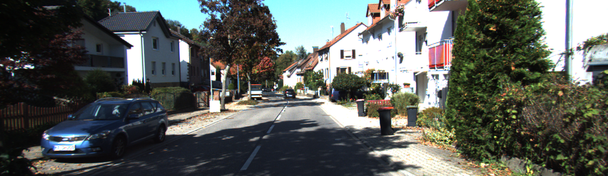}\vspace{0.5ex}
\end{minipage}
}
\subfigure[Appearance flow]{\label{fig:af_multi}
\begin{minipage}[t]{0.22\textwidth}
\includegraphics[width=1.0\textwidth,height=0.35\columnwidth]{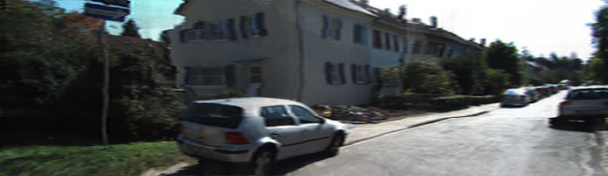}\vspace{0.5ex}
\includegraphics[width=1.0\textwidth,height=0.35\columnwidth]{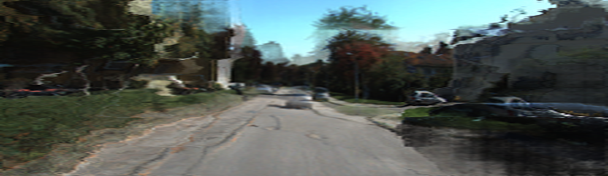}\vspace{0.5ex}
\includegraphics[width=1.0\textwidth,height=0.35\columnwidth]{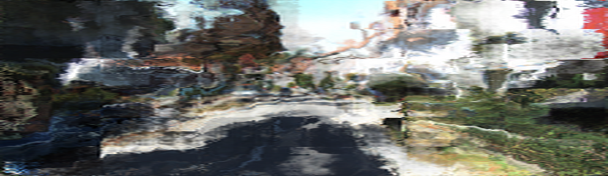}\vspace{0.5ex}
\end{minipage}
}
\subfigure[Our networks (max selection)]{\label{fig:max_multi}
\begin{minipage}[t]{0.22\textwidth}
\includegraphics[width=1.0\textwidth,height=0.35\columnwidth]{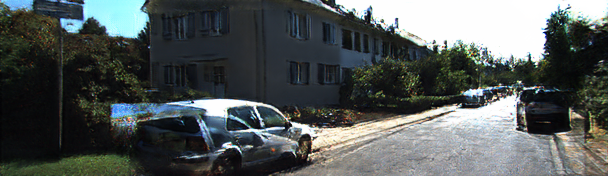}\vspace{0.5ex}
\includegraphics[width=1.0\textwidth,height=0.35\columnwidth]{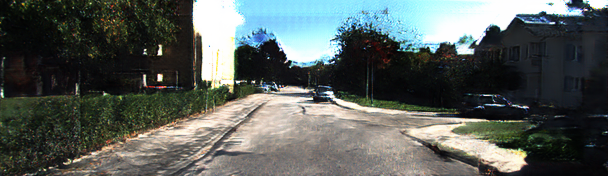}\vspace{0.5ex}

\includegraphics[width=1.0\textwidth,height=0.35\columnwidth]{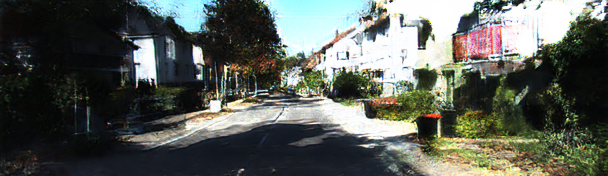}\vspace{0.5ex}

\end{minipage}
}
\subfigure[Our networks (max selection+fusion)]{\label{fig:fusion_multi}
\begin{minipage}[t]{0.22\textwidth}
\includegraphics[width=1.0\textwidth,height=0.35\columnwidth]{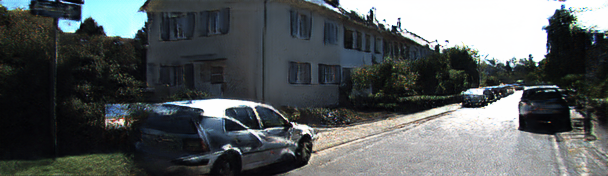}\vspace{0.5ex}
\includegraphics[width=1.0\textwidth,height=0.35\columnwidth]{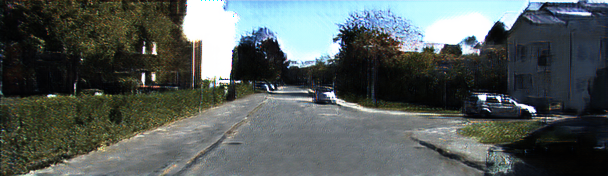}\vspace{0.5ex}
\includegraphics[width=1.0\textwidth,height=0.35\columnwidth]{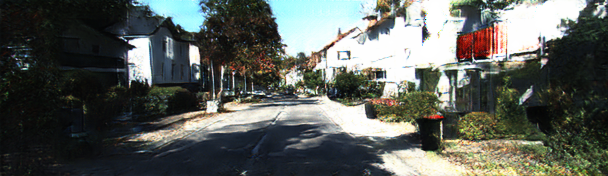}\vspace{0.5ex}
\end{minipage}
}
\caption{Examples of novel view synthesis results with two input images from the KITTI dataset. For each frame, we show (a) ground truth images, (b) multiple image synthesis results generated with appearance flow (c) results produced by max selection in the feature maps of networks, whose parameters are obtained from view synthesis from a single input, and (d) results generated by adding fusion layer after the max selection operator.}
\label{fig:multiple_image_fusion}
\end{figure*}

A single image may not provide enough information to synthesize the image from a novel viewpoint. Bunch of input images from different viewpoints will improve the quality and deterministic of generated image apparently.

We extend our method by adding max selection layer after the sampling layers. The maximum value in the feature map of corresponding position is selected for further layer. From our observation, even without fine-tuning process, we can obtain the fused images. However, the fused images from multiple viewpoints will cause double edges of certain objects. This effect may be caused by the bilinear sampling layers in the down-sampled feature maps. Therefore, another convolutional layer is applied as fusion layer to eliminate this effect.

Both our method and the appearance flow method can take multiple images as inputs. Results of view synthesis from two input images are shown in Figure \ref{fig:multiple_image_fusion}. Two random images from ten adjacent images are selected as inputs to our networks. Firstly, we directly select maximum values in two feature maps as inputs to the following layers. Parameters of this networks are obtained for view synthesis from a single input without fine-tuning process. The synthesized images are shown in Figure \ref{fig:max_multi}. Experiments show that the generated images can keep the main structure of target images. The fusion layer (Conv+BatchNorm+Relu) is introduced as post-processing stage after each max selection operator. After fine-tuning the whole networks, the fusion layer can improve the image quality by enhancing the details and removing the noise pixels. Comparing with our method, images generated with appearance flow method will become distorted for large movements. The appearance flow method is not sensitive to small ego-motion. For example, comparing with the input image, there are small changes in the output image (first row in Figure \ref{fig:af_multi}). Car disappeared in our results shown in bottom row of Figure \ref{fig:max_multi} and \ref{fig:fusion_multi}, because it is not shown in one of the input images. 

The quantitative comparison between appearance flow method and ours' for multiple inputs are listed in Table \ref{tab:multi_view_inception_score}. From the $L_1$ error and inception score shown in the table, images generated by our method (without and with fusion layer after max selection) can preserve the structure of environments.

\begin{table}
\begin{center}
\begin{tabular}{l c c }
\toprule
Method & $L_1$ & Inception Score \\
\midrule
Appearance Flow \cite{Zhou2016View}  & 0.2698 & 3.0959$\pm$0.0788\\
Our networks (max selection)  & 0.2481 & \textbf{3.1409$\pm$0.0621}
 \\
 Our networks & \multirow{2}*{\textbf{0.2318}} & \multirow{2}*{3.0945$\pm$0.1060}\\
 (max selection+fusion)  &  &  \\
\bottomrule
\end{tabular}
\end{center}\caption{Mean pixel $L_1$ error and inception score on the KITTI dataset for novel view synthesis with two input images. The two inputs are randomly selected from ten adjacent frames.}
\label{tab:multi_view_inception_score}
\end{table}

\section{Conclusions}
In this paper, we propose a novel end-to-end framework to generate a image from novel viewpoint. Images are synthesized by convolutional neural networks with generative adversarial loss. Structural information is explicitly incorporated into the feature maps in our networks. Therefore, we can synthesize high-quality images by providing the input image and corresponding viewpoint without distortion for large-scale scenes.  Our framework can be easily extended to the fusion of multiple input images. 

In future work, we would like to incorporate our model with other object moving prediction methods. Stacked generative adversarial networks may be applied to generate high resolution images.

% \clearpage\mbox{}Page \thepage\ of the manuscript.
% \clearpage\mbox{}Page \thepage\ of the manuscript.
% \clearpage\mbox{}Page \thepage\ of the manuscript.
% \clearpage\mbox{}Page \thepage\ of the manuscript.
% \clearpage\mbox{}Page \thepage\ of the manuscript.
% \clearpage\mbox{}Page \thepage\ of the manuscript.
% \clearpage\mbox{}Page \thepage\ of the manuscript.
% This is the last page of the manuscript.
% Now we have reached the maximum size of the ECCV 2018 submission (excluding references).
% References should start immediately after the main text, but can continue on p.15 if needed.

% \clearpage

\bibliographystyle{splncs}
\bibliography{egbib,ref}
\end{document}